\documentclass[10pt,twocolumn,letterpaper]{article}

\usepackage{iccv}
\usepackage{times}
\usepackage{epsfig}
\usepackage{graphicx}
\usepackage{amsmath}
\usepackage{amssymb}
\usepackage{mathtools}
\usepackage{amsthm}
\usepackage{booktabs}
\usepackage{xspace}
\usepackage{multirow,tabularx}
\usepackage{colortbl}
\usepackage[table]{xcolor}
\usepackage{breqn}
\usepackage{subcaption}

\usepackage[textsize=tiny]{todonotes}
\usepackage{enumitem}

\usepackage[breaklinks=true,bookmarks=false]{hyperref}
\usepackage[capitalize, noabbrev]{cleveref}

\crefname{section}{Sec.}{Secs.}
\Crefname{section}{Section}{Sections}
\Crefname{table}{Table}{Tables}
\crefname{table}{Tab.}{Tabs.}

\iccvfinalcopy %

\def\iccvPaperID{29} %

\ificcvfinal\fi

\begin{document}

\newcommand{\spaceholder}{Skill Transformer\xspace}
\newcommand{\method}{Skill Transformer\xspace}
\newcommand{\Reasy}{Rearrange-Easy\xspace}
\newcommand{\Rhard}{Rearrange-Hard\xspace}
\newcommand{\reasy}{rearrange-easy\xspace}
\newcommand{\rhard}{rearrange-hard\xspace}
\newcommand{\miss}{\textcolor{red}{$X\%$}\xspace}
\newcommand{\skillpred}{skill predictor\xspace}

\newcommand{\asn}[1]{\todo[color=red!20, size=\tiny]{AS: #1}}
\newcommand{\as}[1]{\textcolor{red}{AS: #1}}
\newcommand{\xhn}[1]{\todo[color=blue!20, size=\tiny]{XH: #1}}
\newcommand{\xh}[1]{\textcolor{blue}{Haytham: #1}}
\newcommand{\aksn}[1]{\todo[color=orange, size=\tiny]{Akshara: #1}}
\newcommand{\aks}[1]{\textcolor{orange}{Akshara: #1}}

\newcommand{\mono}{Mono\xspace}
\newcommand{\dt}{DT\xspace}
\newcommand{\dtskill}{DT (Skill)\xspace}
\newcommand{\mthree}{M3\xspace}
\newcommand{\mthreeo}{M3 (Oracle)\xspace}
\newcommand{\bcmodular}{BC-Modular\xspace}
\newcommand{\bcmodularo}{BC-Modular (Oracle)\xspace}
\newcommand{\tpsrl}{TP-SRL\xspace}
\newcommand{\skillencoder}{skill inference\xspace}
\newcommand{\Skillencoder}{Skill Inference\xspace}
\newcommand{\actioninfer}{action inference\xspace}
\newcommand{\Actioninfer}{Action Inference\xspace}

\definecolor{Gray}{gray}{0.5}
\definecolor{NewBlue}{rgb}{0.95, 0.95, 1.0}

\title{Skill Transformer: A Monolithic Policy for Mobile Manipulation}

\author{
Xiaoyu Huang\\
Georgia Tech\\
\and
Dhruv Batra\\
Meta AI (FAIR), Georgia Tech\\
\and
Akshara Rai\\
Meta AI (FAIR)\\
\and
Andrew Szot\\
Georgia Tech\\
}

\maketitle
\ificcvfinal\thispagestyle{empty}\fi

\begin{abstract}
We present \method, an approach for solving long-horizon robotic tasks by combining conditional sequence modeling and skill modularity. Conditioned on egocentric and proprioceptive observations of a robot, \method is trained end-to-end to predict both a high-level skill (e.g., navigation, picking, placing), and a whole-body low-level action (e.g., base and arm motion), using a transformer architecture and demonstration trajectories that solve the full task. It retains the composability and modularity of the overall task through a \skillpred module while reasoning about low-level actions and avoiding hand-off errors, common in modular approaches. We test \method on an embodied rearrangement benchmark and find it performs robust task planning and low-level control in new scenarios, achieving a 2.5x higher success rate than baselines in hard rearrangement problems.
\end{abstract}

\section{Introduction}
\label{sec:Introduction}

\begin{figure*}[!h]
  \centering
  \includegraphics[width=0.95\textwidth]{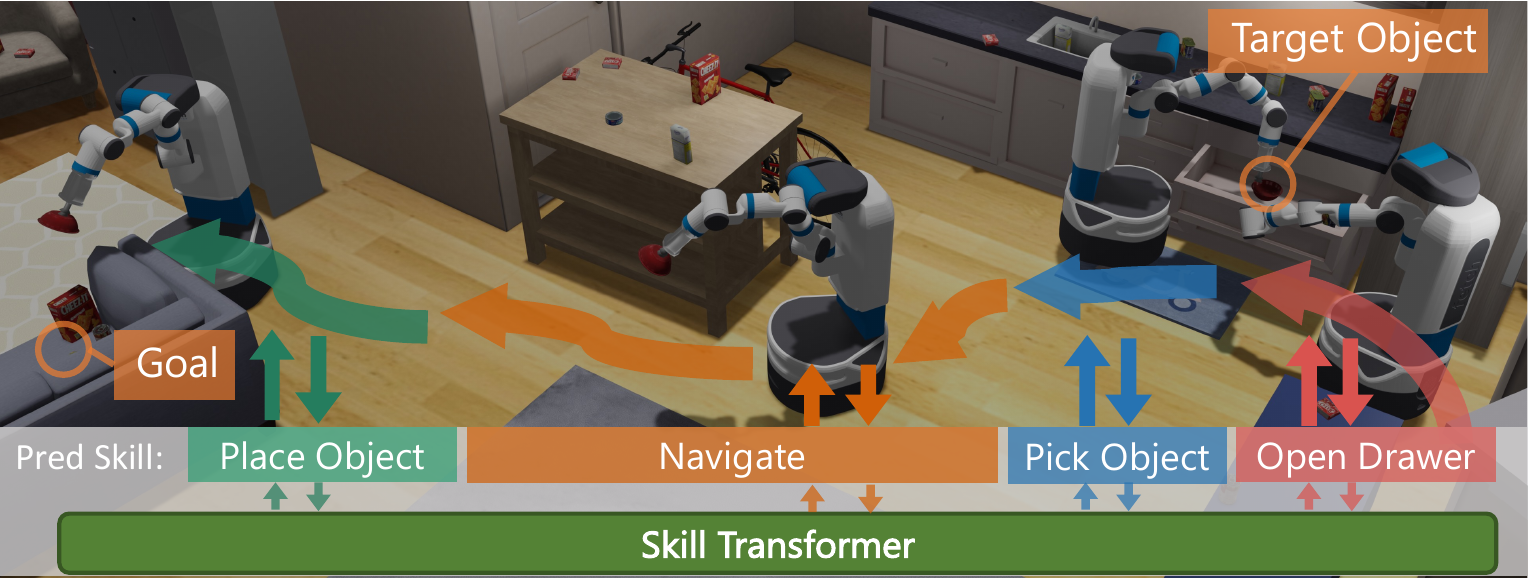}
  \caption{
    A Fetch robot is placed in an unseen house and must rearrange an object from a specified start to a desired goal location from egocentric visual observations. The robot infers the drawer is closed and opens the drawer to access the object. Next, it picks the object, navigates to the goal, and places the object. The Skill Transformer is a monolithic ``pixels-to-actions" policy for such long-horizon mobile manipulation problems which infers the active skill and the low-level continuous control.
    }
  \label{fig:teaser}
  \vspace{-10pt}
\end{figure*}

A long-standing goal of embodied AI is to build agents that can seamlessly operate in a home to accomplish tasks like preparing a meal, tidying up, or loading the dishwasher. To successfully carry out such tasks, an agent must be able to perform a wide range of diverse skills, like navigation and object manipulation, as well as high-level decision-making, all while operating with limited egocentric inputs. Our particular focus is the problem of embodied object rearrangement~\cite{habitatrearrangechallenge2022}. An example in \Cref{fig:teaser} shows that the Fetch mobile manipulator must rearrange a bowl from an initial position to a desired goal position (both circled in orange) in an unfamiliar environment without access to any pre-existing maps, 3D models, or precise object locations.
Successfully completing this task requires executing both low-level motor commands for accurate arm and base movements, as well as high-level decision-making, such as determining whether a drawer needs to be opened before accessing a specific object.

Long-horizon tasks like object rearrangement require many low-level control steps and naturally break down into distinct phases of skills such as navigation, picking, placing, opening, and closing. 
Prior works~\cite{brohan2022rt,reed2022generalist,shridhar2022cliport} show that it is possible to end-to-end learn a single policy that can perform such diverse tasks and behaviors. 
Such a monolithic policy directly maps observations to low-level actions, reasoning both about which skill to execute and how to execute it.
However, scaling end-to-end learning to long-horizon, multi-phase tasks with thousands of low-level steps and egocentric visual observations remains a challenging research question. 
Other works decompose the task into individual skills and then separately sequence them to solve the full task~\cite{szot2021habitat,gu2022multi}.
While this decomposition makes the problem tractable, this modularity comes at a cost. 
Sequencing skills suffers from hand-off errors between skills, where one skill ends in a situation where the next skill is no longer feasible. 
Moreover, it does not address how to learn the skill sequencing or how to adapt the planned sequence to unknown disturbances. 

We present \textbf{Skill Transformer}, an end-to-end trainable transformer policy for multi-skill tasks. \method predicts both low-level actions and high-level skills, conditioned on a history of high-dimensional egocentric camera observations. 
By combining conditional sequence modeling with a skill prediction module, \method can model long sequences of both high and low-level actions without breaking the problem into disconnected sub-modules. In \Cref{fig:teaser}, \method learns a policy that can perform diverse motor skills, as well as high-level reasoning over which motor skills to execute. We train \method autoregressively with trajectories that solve the full rearrangement task. \method is then tested in unseen settings that require a variable number of high-level and low-level actions, overcoming unknown perturbations.

\method achieves state-of-the-art performance in rearrangement tasks requiring variable-length high-level planning and delicate low-level actions, such as rearranging an object that can be in an open or closed receptacle, in the Habitat environment~\cite{habitatrearrangechallenge2022}. 
It achieves 1.45x higher success rates in the full task and up to 2.5x higher in especially challenging situations over modular and end-to-end baselines and is more robust to environment disturbances, such as a drawer snapping shut after opening it, despite never seeing such disturbances during training. Furthermore, we demonstrate the importance of the design decisions in \method and its ability to scale with larger training dataset sizes.
The code is available at \href{https://bit.ly/3qH2QQK}{https://bit.ly/3qH2QQK}.

\section{Related Work}
\label{sec:Relatedworks}

\textbf{Object Rearrangement}: Object rearrangement can be formulated as a task and motion planning~\cite{garrett2021integrated} problem where an agent must move objects to a desired configuration, potentially under some constraints. We refer to \cite{batra2020rearrangement} for a comprehensive survey. Prior works have also used end-to-end RL training for learning rearrangement~\cite{weihs2021visual, ramrakhya2022habitat, gan2021threedworld, ehsani2021manipulathor}, but unlike our work, they rely on abstract interactions with the environment, no physics, or simplified action spaces. Other works focus on tabletop rearrangement~\cite{liang2022code,qt-opt,florence2022implicit} while our work focuses on longer horizon mobile manipulation. Although mobile manipulation has witnessed significant achievements in robotics learning \cite{minniti2019whole, wang2020learning, li2020hrl4in, mittal2022articulated, honerkamp2021learning}, due to the vastly increased complexity, in rearrangement tasks, a majority of previous works \cite{brohan2022rt, szot2021habitat} explicitly separate the base and arm movements to simplify the task. A few works \cite{gu2022multi, wong2022error, wijmansver} have shown the superiority of mobile manipulation over static manipulation. In the Habitat Rearrangement Challenge~\cite{szot2021habitat,habitatrearrangechallenge2022}, methods that decompose the rearrangement task into skills are the most successful~\cite{gu2022multi,wijmansver} but they require a fixed high-level plan and cannot dynamically select skills based on observations during execution.

\textbf{Transformers for Control}: Prior work uses a transformer policy architecture to approach RL as a sequence modeling problem~\cite{chen2021decision,zheng2022online,janner2021reinforcement,carroll2022towards}. Our work also uses a transformer-based policy, but with a modified architecture that predicts high-level skills along with low-level actions. Other works demonstrate that transformer-based policies are competent multi-task learners~\cite{lee2022multi,xu2022prompting,shridhar2022perceiver,brohan2022rt,reed2022generalist}. Likewise, we examine how a transformer-based policy can learn multiple skills and compose them to solve long-horizon rearrangement tasks. \cite{liang2022transformer,bonatti2022pact} further show that transformers can encode underlying dynamics of the environment that benefits learning for individual skills. 

\textbf{Skill Sequencing}: Other works investigate how to sequence together skills to accomplish a long horizon task. By utilizing skills, agents can reason over longer time horizons by operating in the lifted action space of skills rather than over low-level actions~\cite{sutton1999between}. Skills may be either learned from demonstrations~\cite{fox2017multi,ajay2020opal,hakhamaneshi2021hierarchical}, from trial and error learning~\cite{sharma2019dynamics,eysenbach2018diversity}, or hand defined~\cite{lee2018composing}. This work demonstrates that a single transformer policy can learn multiple skills. However, when sequencing together skills, there may be ``hand-off" issues when skills fail to transition smoothly. Prior works tackle this issue by learning to match the initiation and termination sets of skills~\cite{lee2021adversarial,clegg2018learning} or learning new skills to handle the chaining~\cite{bagaria2019option,konidaris2009skill}. 
Decision Diffuser~\cite{ajay2022conditional} learns a generative model to combine skills, but requires specifying the skill combination order.
We overcome these issues by distilling all skills into a single policy and training this monolithic policy end-to-end to solve the overall task while maintaining skill modularity through a skill-prediction module.

\section{Task}
\label{sec:task} 

We focus on the problem of embodied object rearrangement, where a robot is spawned in an unseen house and must rearrange an object from a specified start position to a desired goal position entirely from on-board sensing and without any pre-built maps, object 3D models, or other privileged information. We follow the setup of the 2022 Habitat Rearrangement Challenge~\cite{habitatrearrangechallenge2022}. Specifically, a simulated Fetch robot ~\cite{fetchrobot} senses the world through a $ 256 \times 256$ head-mounted RGB-D camera, robot joint positions, whether the robot is holding an object or not, and base egomotion giving the relative position of the robot since the start of the episode. The 3D position of the object to rearrange, along with the 3D position of where to rearrange the object to is also provided to the agent.

The agent interacts with the world via a mobile base, 7DoF arm, and suction grip attached to the arm. The episode is successful if the agent calls an episode termination action when the object is within 15cm of its target. The agent fails if it excessively collides with the environment or fails to rearrange the object within the prescribed low-level time steps. 
See \Cref{sec:further-task-details} for more task details.

We evaluate our method in the \textbf{\Rhard Task}, which presents challenges in both low-level actions and high-level reasoning. Unlike the Habitat Rearrangement tasks~\cite{gu2022multi, szot2021habitat} that prior works focus on, this task involves objects and goals that may spawn in closed receptacles. Therefore, the agent may need to perform intermediate actions, such as opening a drawer or fridge to access the target object or goal. Furthermore, the correct sequence of skills required to solve the task differs across episodes and is unavailable to the agent. As a result, the agent must infer the correct behaviors from egocentric visual observations and proprioception. This more natural setting better reflects real-world rearrangement scenarios, thus making it the focus of this work. 

From the dataset provided in \cite{habitatrearrangechallenge2022}, we create a harder version of the full Rearrange track dataset for evaluation. In this harder version, more episodes have objects and goals contained in closed receptacles. Specifically, the updated episode distribution in evaluation for the \rhard task is divided into three splits: 
\begin{itemize}[itemsep=0pt,topsep=0pt,parsep=0pt,partopsep=0pt,parsep=0pt,leftmargin=*]
\item 25\% \emph{easy} scenes with objects initially accessible.
\item 50\% \emph{hard} scenes with objects in closed receptacles.
\item 25\% \emph{very hard} scenes with both objects and goals in closed receptacles. 
\end{itemize}
The original \rhard dataset is used for training.

\begin{figure*}[!h]
  \centering
  \includegraphics[width=0.9\textwidth]{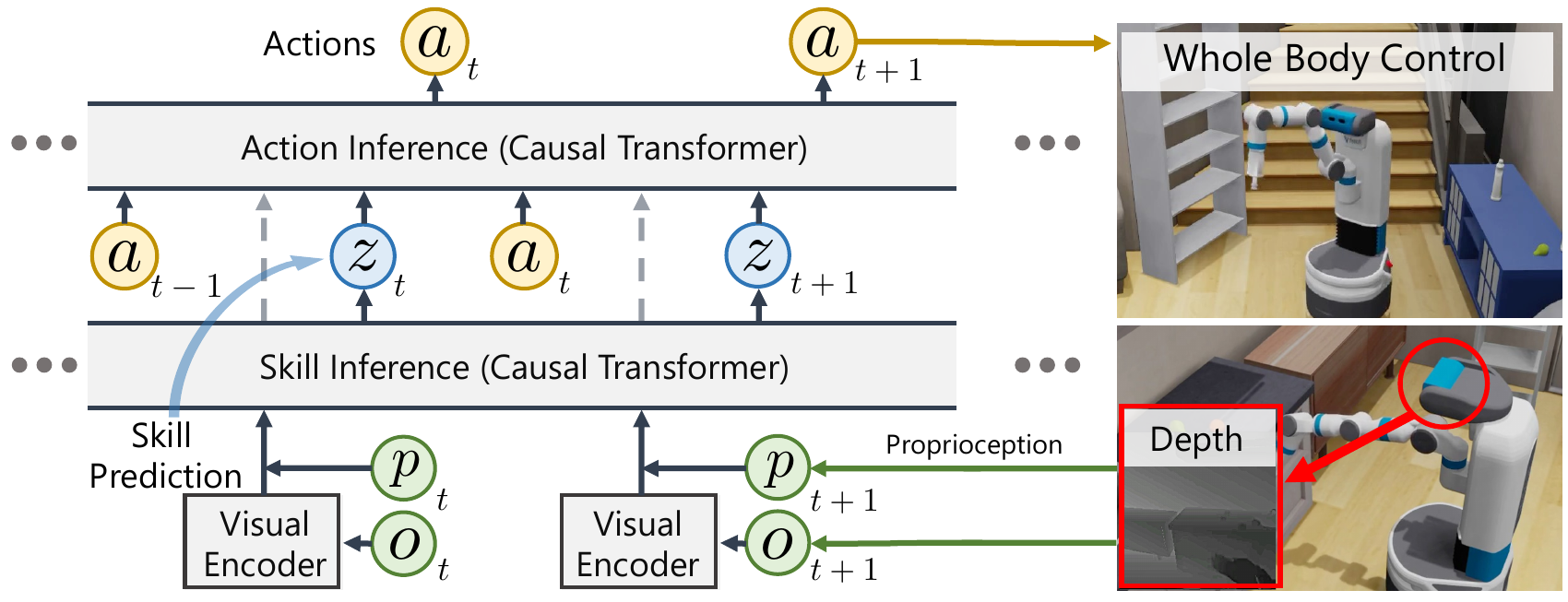}
  \vspace{-5pt}
  \caption{
    The \method architecture. \method learns to map egocentric depth images from the Fetch robot's head camera and proprioceptive state to actions controlling the robot, including the base movement, arm movement, and gripper state. First, the raw visual observations are processed by a visual encoder and concatenated with the proprioceptive state. Next, the \Skillencoder module, implemented as a causal transformer sequence model, infers the skill ID that is executing for each time step. Finally, the \Actioninfer module predicts the low-level action to execute on the robot from the inferred skill, observation, and previous action information. The model is trained through end-to-end imitation learning.
    }
  \label{fig:method}
  \vspace{-10pt}
\end{figure*}

\section{Method}
\label{sec:Method}

We propose \emph{\method}, an approach for learning long-horizon, complex tasks that require sequencing multiple distinct skills. In \Cref{sec:method-arch}, we describe the components of \method, which include a Skill Inference module for reasoning over skill selections and an Action Inference module to predict actions for controlling the robot, both modeled using a Transformer architecture and conditioned on past and current observations. Next, in \Cref{sec:method-training}, we detail how we train \method in an autoregressive fashion from demonstrations of the full task.

\subsection{\method Architecture}
\label{sec:method-arch} 

\method breaks the long-horizon task of object rearrangement into first inferring skill selections using a Skill Inference module and then predicting low-level control using an Action Inference module, given this skill choice. Both modules are implemented as causal transformer neural networks~\cite{radford2018improving}, conditioned on past and current observations of the robot. 

\textbf{Input:} Both modules in \method take as input a sequence of tokens that encode robot observations. We refer to the maximum length of the input sequence as the context length $C$. Let $ o_t $ denote the $ 256 \times 256 \times 1$ depth visual observations, and $ p_t$ denote the proprioceptive inputs of the joint angles and holding state at time $ t$. The visual observation $ o_t$ is first encoded with a Resnet-18 encoder~\cite{he2016deep} to a $256$-dimension visual embedding. Likewise, non-visual inputs are encoded with a multi-layer perceptron network into a $ 256$-dimension sensor embedding. The visual and non-visual continuous embeddings are concatenated to form a $ 512$-dimension input token $ x_t$. We also add a per-time step positional embedding to each input token $ x_t$ to enable modeling time across the sequence. $x_t$ forms the input token that encodes the current observations, and \method takes as input a history of such tokens up to context length $C$. Note that we do not use RGB input as prior work~\cite{szot2021habitat} found it unnecessary for geometric object rearrangement, and training with only depth input is faster. 

\textbf{\Skillencoder:} The first module of \method is the \Skillencoder module which predicts the current skill to be executed. The \Skillencoder module takes as input a sequence of the input embeddings $ \left\{ x_{t-C}, \dots , x_{t} \right\} $ through a causal transformer network~\cite{vaswani2017attention} with a $ 512$ hidden dimension, to produce a one-hot skill vector $ c_t$ corresponding to the predicted skill that should be active at the current time step $ t$. Using the causal mask, label $ c_{i}$ only has access to the previous tokens $ \left\{ t : t < i \right\} $ in the input sequence. Then, we embed $ c_{i}$ into a latent skill embedding $ \left\{ z_{t-C}, \dots , z_{t} \right\} $.  

\textbf{\Actioninfer:} Next, the \Actioninfer module outputs a distribution over robot low-level actions. The \Actioninfer module takes as input the input embedding $ x_t $, an embedding of the previous time step action  $ a_{t-1}$, and the skill embedding $ z_t$. Each of these inputs are represented as distinct tokens in the input sequence of $\left\{ x_{t-C}, a_{t-C-1}, z_{t-C}, \dots, x_{t}, a_{t-1}, z_{t} \right\}$. This sequence is then fed into a transformer with the same architecture as the \Skillencoder module. By taking as input the skill embeddings $ z_t$, the \Actioninfer module learns consistency in executing individual skills and the transitions between them. As we will show in \Cref{sec:ablation}, this also allows training \method with datasets that only successfully complete a portion of the skills.

The \Actioninfer module then outputs a sequence of predicted next actions $ \left\{ a_{t-C}, \dots a_{t} \right\} $ which control the robot. These actions determine the whole body control actions composed of three components: arm joint positions, base velocities, and gripper state. The module first outputs a $ 12$-dimension output vector. The first seven components are used as the mean of a normal distribution to sample the desired delta joint positions, which will then be actuated via a PD controller. The next two dimensions are used as the mean of a normal distribution to sample the desired base velocities, scaled by the maximum linear and angular velocities of the robot base. The remaining three outputs are passed through a softmax layer to generate probabilities for the gripper action, which indicate opening,  closing, or maintaining the previous gripper state. 

Unless otherwise specified, we use 2 transformer layers and a context length $C=30$ for all architectures, giving 2M parameters for the transformer network in the \Skillencoder module. The \Actioninfer module is also trained with the same architecture and a context length $C$ of 30 time steps, translating to a sequence length of 90 tokens. In total, between the observation encoders, \Skillencoder, and \Actioninfer modules, the \method policy network consists of 16M parameters.

\subsection{\method Training}
\label{sec:method-training}
In this section, we describe how \method is trained end-to-end to learn long-horizon rearrangement tasks from imitation learning. We train the policy using supervised learning on demonstration trajectories produced by a data-generating policy. We assume the data-generating policy demonstrates a composition of skills to solve the task. Such compositional approaches are the state-of-the-art for the Habitat Rearrangement task~\cite{szot2021habitat,gu2022multi}.

Specifically, we assume the data-generating policy demonstrates one of $ N$ skills, and each time step $t$ is labeled with the skill ID $y_t^{\text{skill}}$ that is currently executing. In our problem setting, these skills are: pick, place, navigate, open drawer, and open fridge. We also record the policy's actions at time step $t$ that consists of arm action $y_{t}^{\text{a}}$, base velocity $y_{t}^{\text{b}}$, and gripper state $y_{t}^{\text{g}}$. In addition, we acquire from the simulator a one-dimensional indicator label $y_t^{\text{pred}}$ defined as $y_t^{\text{pred}} = 0$ if the object does not start in a closed receptacle and $y_t^{\text{pred}} = i$ if $i^{\text{th}}$ receptacle contains the object, at the time step $t$. 

\setlength{\belowdisplayskip}{0pt} \setlength{\belowdisplayshortskip}{0pt}
\setlength{\abovedisplayskip}{0pt} \setlength{\abovedisplayshortskip}{0pt}

\subsubsection{Training Loss}
We separately supervise the \Skillencoder and \Actioninfer modules. We supervise the \Skillencoder module over the context length $C$ using loss function: 
\begin{align}
  \label{eq:skill-loss}
  \mathcal{L}_{\scriptscriptstyle \text{skill}} = \sum_{t=1}^C \mathcal{L}_{\scriptscriptstyle  \text{FL}}(c_{t}, y_t^{\scriptscriptstyle \text{skill}})
\end{align} 
where $\mathcal{L}_{\text{FL}}$ represents the focal loss. The focal loss counteracts the imbalance of skill execution frequencies. For example, the navigation skill executes more frequently than the other skills as it is needed multiple times within a single episode, and takes more low-level time steps to execute than manipulation skills.

In order to train the \Actioninfer module in parallel with the \Skillencoder module, we use teacher-forcing, and input the ground truth skill labels during training, instead of the inferred distributions that are used at test time. We supervise the \Actioninfer module over the context length $C$ with loss function: 
\begin{equation}\label{eq:action-loss}
\begin{split}
  \mathcal{L}_{\scriptscriptstyle \text{act}} = \sum_{t=1}^C (&\mathcal{L}_{\scriptscriptstyle \text{MSE}}(a_{t}^{\text{b}}, y_{t}^{\text{b}}) +\mathcal{L}_{\scriptscriptstyle \text{MSE}}(a_{t}^{\text{a}}, y_{t}^{\text{a}}) 
  + \mathcal{L}_{\scriptscriptstyle \text{FL}}(a_{t}^{\text{g}}, y_{t}^{\text{g}})  )
\end{split}
\end{equation}
where $\mathcal{L}_{\text{MSE}}$ represents the mean-squared error loss.

In addition, we introduce auxiliary heads and losses for the \Skillencoder module. The auxiliary head reconstructs oracle information unavailable to the policy. Specifically, assuming there exist $M$ receptacles that may contain the target object, the auxiliary head outputs an $M+1$ dimensional distribution $v_t$. We then supervise $v_t$ with $y_t^{\text{pred}}$ over the context length $C$ via the loss function: 
\begin{equation}~\label{eq:aux-loss}
  \mathcal{L}_{\scriptscriptstyle \text{aux}} = \sum_{t=1}^C \mathcal{L}_{\scriptscriptstyle \text{CE}}(v_{t}, y_t^{\scriptscriptstyle \text{pred}})
\end{equation}
where $\mathcal{L}_{\text{CE}}$ represents the Cross-Entropy loss. 

We incorporate this auxiliary loss because the data-generating policy used to generate the demonstrations has a low success rate in correctly opening the receptacles. This low success rate can hinder learning because \method needs to reason if it failed due to not opening the receptacle at all or opening an incorrect receptacle. The auxiliary loss incentivizes \method to identify the key features that lead to the decision to open specific receptacles and to avoid wrong ones. This auxiliary objective with privileged information is only used for training and not used at test time.

The final loss is the unweighted sum of individual loss terms: $ \mathcal{L} = \mathcal{L}_{\text{skill}} + \mathcal{L}_{\text{action}} + \mathcal{L}_{\text{aux}}$. Note that in training, we calculate the loss over all time steps, whereas at test time, we only output action $a_t$ for the current time step $t$ to the environment. 
While training, we also train the visual encoder, in contrast to previous works using transformers in embodied AI that only demonstrate performance with pre-trained visual encoders~\cite{chen2021decision, brohan2022rt}.
Since \method is an end-to-end trainable policy, it can also be trained with offline-RL using the rewards from the dataset or with online-RL by interacting with the simulator. 
We leave exploring \method with RL training for future work.

\subsubsection{Training Scheme}
We use a similar scheme as \cite{chen2021decision}, where we input a sequence of 30 timesteps and compute the loss for the skill and action distributions for all timesteps in the sequence using autoregressive masking. 
\method uses a learning warm-up phase where the learning rate is linearly increased at the start of training.
After the warm-up phase, \method uses a cosine learning rate scheduler which decays to 0 at the end of training. 
The entire dataset was iterated through approximately 6 times in total during the complete training process. We report the performance of the checkpoint with the highest success rate on seen scenes. 
We provide further details about \method in \Cref{sec:supp:method}.

\section{Experiments}
\label{sec:Experiments}

In this section, we describe a series of experiments used to validate our proposed approach. 
First, we introduce the experiment setup and state-of-the-art baselines for comparison. Then, we quantify the performance of \method in the \rhard task and compare it to the baselines. 
Furthermore, the robustness of the methods to disturbances at evaluation time are evaluated by adversarial perturbations, like closing drawers after the agent has opened them.  
Finally, we perform an ablation study to confirm the key design choices involved in \method.

\subsection{Experiment Setup}
\label{sec:exp-setup}

We conduct all experiments in the Habitat 2.0 \cite{szot2021habitat} simulation environment and the rearrangement tasks as described in \Cref{sec:task}. For training, we use the \emph{train} dataset split from the challenge consisting of 50,000 object configurations across 63 scenes. Each episode consists of starting object positions and a target location for the object to be rearranged. The robot positions are randomly generated in each episode. We first train individual skills (pick, place, navigate, open, close) that are used by a data generating policy to generate 10,000 demonstrations for training \method. To train skills, we use the Multi-skill Mobile Manipulation (M3)~\cite{gu2022multi} method with an oracle task planner with privileged information about the environment. This approach trains skills with RL and sequences them with an oracle planner. Further details of the data generating policy and the demonstration collected are detailed in \Cref{sec:expert-details}.

\subsection{Baselines}
\label{sec:exp-baseline}

We compare \method (ST) to the following baselines in the \rhard task. The parameter count of each baseline policy is adjusted to match the 16M parameters in \method.
\begin{itemize}[itemsep=0pt,topsep=0pt,parsep=0pt,partopsep=0pt,parsep=0pt,leftmargin=*]
  \item Monolithic RL (\textbf{Mono}): Train a monolithic neural network directly mapping sensor inputs to actions with end-to-end RL using DD-PPO~\cite{wijmans2019dd}, a distributed form of PPO~\cite{schulman2017proximal}. The monolithic RL policy is trained for 100M timesteps, reflecting a wall time of 21h. 

  \item Decision Transformer (\textbf{DT}) ~\cite{chen2021decision}: Train the monolithic policy from the offline dataset using a decision transformer architecture. The DT policy is trained with the same learning rate scheme and number of epochs as \method. At test time, we condition the \dt policy with the highest possible reward in the dataset. 

  \item Decision Transformer with Auxiliary Skill Loss (\textbf{DT (Skill)}): Train the Decision Transformer with an auxiliary loss of skill prediction. This baseline incorporates an auxiliary head that predicts the current skill during training and is trained with focal loss like \method. 
\end{itemize}
Note that Decision Transformer requires the environment reward at evaluation time to compute the reward-to-go input. \method and the other baselines do not require this assumption. We also compare to the following hierarchical baselines which first decompose the task into skills and then sequence these skills to solve the tasks.
\begin{itemize}[itemsep=0pt,topsep=0pt,parsep=0pt,partopsep=0pt,parsep=0pt,leftmargin=*]
  \item Task-Planning + Skills RL (\textbf{TP-SRL})~\cite{szot2021habitat}: Individually train pick, place, navigate, and open skills using RL and then sequence the skills together with a fixed task planner. Each skill predicts when it should end and the next skill should begin by a skill termination action. TP-SRL trains all manipulation skills with a fixed base.

  \item Multi-skill Mobile Manipulation \textbf{(M3)}~\cite{gu2022multi}: \mthree is a modular approach similar to \tpsrl, but instead trains skills as \emph{mobile} manipulation skills with RL. All these skills are trained to be robust to diverse starting states. Like TP-SRL, it uses a fixed task planner to sequence the skills. 

  \item Behavior Cloning Modular skills \textbf{(BC-Modular)}: Individually train transformer-based policies for each of the skills using behavior cloning with the same training dataset as \method. Then sequence together with the same fixed task planner as the other modular baselines. This baseline helps measure the gap in performance from training policies with behavior cloning as opposed to the RL-trained policies, \mthree and \tpsrl.
\end{itemize}

However, these modular baselines are designed to only execute a fixed pre-defined sequence of skills (navigate, pick, navigate, place). They cannot adjust their skill sequence based on environment observations, i.e. the task plan is inaccurate when the robot needs to open a receptacle to pick or place an object. 

To address this, we additionally include a version of the modular baselines, \textbf{M3 (Oracle)} and \textbf{BC-Modular (Oracle)}, that extracts the correct task plan directly from the simulator, which we refer to as an \emph{oracle} plan. In contrast, the proposed method, \method, infers which skill to execute from egocentric observations without using this privileged oracle plan. Note that neither modular baseline re-executes skills if they fail, while \method can re-plan. For further details about baselines and all hyperparameters, see \Cref{sec:baseline-details}

\subsection{\Rhard}
\label{sec:exp-rhard}
First, we show the performance of \method on the \rhard task described in \Cref{sec:task}. In this task, the agent must rearrange a single object from the start to the goal, and the object or the goal may be located in a closed receptacle. Therefore, methods need high-level reasoning to decide if the receptacle should first be opened or not. 

We evaluate the success rate of all methods on both ``Train" (Seen) and ``Eval" (Unseen) scenes and object configurations. The success rate is measured as the average number of episodes where the agent placed the target object within 15cm of the goal within 5000 steps and without excessively colliding with the environment.
Specifically, the episode will terminate with failure if the agent accumulates more than 100kN of force throughout the episode or 10kN of instantaneous force due to collisions with the scene or objects.
We also report the success rate on each of the three dataset splits described in \Cref{sec:task}. 
Our evaluation dataset splits have 100 episodes in the easy split, 200 in the hard split (100 for objects starting in closed receptacles and 100 for goals in closed receptacles), and 100 in the very hard split.

\begin{table*}[h]
  \centering
  \resizebox{1.98 \columnwidth}{!}{
    \begin{tabular}{ccccccccc}
\toprule
& \multicolumn{4}{c}{\textbf{Train} (Seen)} & \multicolumn{4}{c}{\textbf{Eval} (Unseen)} \\
\cmidrule(rl){2-5} \cmidrule(rl){6-9}
Method & \textbf{All Episodes} & \hspace*{1.5mm} \textbf{Easy} \hspace*{1.5mm} & \hspace*{1.5mm} \textbf{Hard} \hspace*{1.5mm} & \textbf{Very Hard} & \textbf{All Episodes}  & \hspace*{1.5mm} \textbf{Easy} \hspace*{1.5mm} & \hspace*{1.5mm} \textbf{Hard} \hspace*{1.5mm} & \textbf{Very Hard}\\
\midrule
\color{Gray} \textbf{BC-Modular (Oracle)} & \color{Gray}  5.67{\scriptsize $\pm$ 0.12} & \color{Gray}  20.33{\scriptsize $\pm$ 1.70} & \color{Gray}  0.83{\scriptsize $\pm$ 0.24}  & \color{Gray}  0.00{\scriptsize $\pm$ 0.00}  & \color{Gray}  5.00{\scriptsize $\pm$ 0.35} & \color{Gray}  15.33{\scriptsize $\pm$ 1.70} & \color{Gray}  1.67{\scriptsize $\pm$ 1.65} & \color{Gray}  0.33{\scriptsize $\pm$ 0.47}   \\
\color{Gray} \textbf{M3 (Oracle)} & \color{Gray}  22.92 {\scriptsize $ \pm$ 0.72}  & \color{Gray}  59.33 {\scriptsize $ \pm $ 1.70}  & \color{Gray}  13.00 {\scriptsize $ \pm $ 1.41}  & \color{Gray}  6.33 {\scriptsize $ \pm $ 0.47} & \color{Gray}  22.58 {\scriptsize $ \pm$ 1.18}   & \color{Gray}  58.67 {\scriptsize $ \pm $ 1.89} & \color{Gray} 12.67 {\scriptsize $ \pm$ 2.36}  & \color{Gray}  6.33 {\scriptsize $ \pm$ 0.47}  \\
\midrule
\textbf{BC-Modular} &  5.08{\scriptsize $\pm$ 0.42}  & 20.33{\scriptsize $\pm$ 1.70} & 0.00{\scriptsize $\pm$ 0.00} & 0.00{\scriptsize $\pm$ 0.00} &  3.83{\scriptsize $\pm$ 0.42} &  15.33{\scriptsize $\pm$ 1.70} &  0.00{\scriptsize $\pm$ 0.00} &  0.00{\scriptsize $\pm$ 0.00}   \\
\textbf{TP-SRL} &  7.25 {\scriptsize $ \pm$ 0.82}  & 29.00{\scriptsize $ \pm$ 3.27} & 0.00 {\scriptsize $ \pm$ 0.00} & 0.00 {\scriptsize $ \pm$ 0.00} &  6.67 {\scriptsize $ \pm$ 0.85}  & 26.67 {\scriptsize $ \pm$ 3.40} & 0.00 {\scriptsize $ \pm$ 0.00} & 0.00 {\scriptsize $ \pm$ 0.00}  \\
\textbf{M3} &  14.00 {\scriptsize $ \pm$ 0.35}  & \textbf{55.67} {\scriptsize $ \pm$ 0.94} & 0.16 {\scriptsize $ \pm$ 0.47} & 0.00 {\scriptsize $ \pm$ 0.00} &  13.25 {\scriptsize $ \pm$ 1.08} & \textbf{53.00} {\scriptsize $ \pm$ 4.32} & 0.00 {\scriptsize $ \pm $ 0.00} & 0.00 {\scriptsize $ \pm$ 0.00}   \\
\textbf{Mono} &  0.00 {\scriptsize $\pm$ 0.00}  &  0.00 {\scriptsize $\pm$ 0.00}  &  0.00 {\scriptsize $\pm$ 0.00} &  0.00 {\scriptsize $\pm$ 0.00} &  0.00 {\scriptsize $\pm$ 0.00} &  0.00 {\scriptsize $\pm$ 0.00} &  0.00 {\scriptsize $\pm$ 0.00} &  0.00 {\scriptsize $\pm$ 0.00} \\
\textbf{DT} &  7.08 {\scriptsize $\pm$ 0.63}  & 19.33 {\scriptsize $\pm$ 0.47} & 3.83 {\scriptsize $\pm$ 0.62} & 2.00 {\scriptsize $\pm$ 0.94} &  6.50 {\scriptsize $\pm$ 0.41}  &  15.33 {\scriptsize $\pm$ 1.25} &  3.83 {\scriptsize $\pm$ 0.62} &  3.00 {\scriptsize $\pm$ 1.63}  \\
\textbf{DT (Skill)} &  8.17 {\scriptsize $\pm$ 0.67}  & 23.33 {\scriptsize $\pm$ 2.49} & 3.00{\scriptsize $\pm$ 2.16} & 3.33{\scriptsize $\pm$ 0.94} &  7.67{\scriptsize $\pm$ 0.66}  &  18.67{\scriptsize $\pm$ 2.62} &  4.83{\scriptsize $\pm$ 1.70} &  2.33{\scriptsize $\pm$ 1.25}  \\
\rowcolor{NewBlue}
\textbf{ST (Ours)} &  \textbf{21.08}{\scriptsize $\pm$ 1.36}  &  44.00{\scriptsize $\pm$ 0.82} &  \textbf{15.67}{\scriptsize $\pm$ 1.70}  &  \textbf{9.00}{\scriptsize $\pm$ 2.16} &  \textbf{19.17}{\scriptsize $\pm$ 1.05}  & 37.00{\scriptsize $\pm$ 0.82} & \textbf{15.83}{\scriptsize $\pm$ 0.85} &\textbf{8.00}{\scriptsize $\pm$ 3.27}  \\
\bottomrule
\end{tabular}

  }
  \caption{
    Success rates on the rearrangement task. ``Eval" results are from test episodes in unseen scenes. In ``Easy", objects and goals start on accessible locations. In ``Hard", objects start in closed receptacles. In ``Very Hard" objects and goals start in closed receptacles. All results are averages across 100 episodes except for the ``Hard" split which is an average across 200 episodes. Grayed methods at the top use the oracle skill order. We report the mean and standard deviation across 3 random training seeds. For hierarchical methods, we retrain all the skills for each random seed.
  }
  \label{tab:hard-main}
\end{table*}

\Cref{tab:hard-main} shows that \method achieves higher success rates in both the ``Train" and ``Eval" settings than all of the baselines that do not have oracle planning. The \mono baseline cannot find any success, even after training for 100M steps. This demonstrates the difficulty of learning the \rhard task from a reward signal alone.

The modular \tpsrl, \mthree, and \bcmodular methods rely on a fixed task plan to execute the separately learned skill policies. As a result, the state-of-the-art \mthree method achieves a significantly lower success rate than \method in the overall result as well as the \emph{hard} and \emph{very hard} splits because it is unable to adapt the skill sequence to different situations. On the other hand, \method does not rely on a fixed ordering of skills and can adapt its high-level task-solving to new situations, achieving a 35\% increase in success rate across all episodes. 

Notice that even with an oracle plan, the success rate of M3 (Oracle) on the overall result for unseen scenes is only 3\% higher than that of \method. More importantly, \method outperforms M3 (Oracle) in both \emph{hard} and \emph{very hard} settings, even without oracle task planning information. We find that \method consistently achieves higher success rates in episodes with hidden objects or goals, while M3 (Oracle) experiences a substantial decrement as the difficulty increases. This difference is likely due to the low success rate in M3 (Oracle)'s opening skill, hindering its ability to complete the skill in one attempt. In contrast, as we will discuss quantitatively later in \Cref{sec:exp-robustness}, \method is capable of detecting and retrying failed skills, eventually leading to better performance in retrieving and placing the object. 

Further observe that \bcmodular exhibits the lowest performance among modular methods, with an overall success rate of only 5\% even when employing oracle task planning. This indicates that policies trained via Behavior Cloning, which uses only one-tenth the experience of those trained with RL, encounter greater difficulty in executing intricate mobile manipulation skills. Conversely, \method, featuring the multi-skill \Actioninfer module, mitigates this limitation and achieves up to 4x performance improvements across all episodes. 

 \dt and \dtskill can also achieve some success in \emph{hard} and \emph{very hard} splits. Nonetheless, the success rates are significantly lower than \method, especially in the \emph{very hard} split. We observe that \dt-based methods lack the capability to execute skills consistently and often mix up different skill behaviors. Also, notice that \dtskill with an auxiliary skill loss does not show a significant improvement in performance compared to \dt without the skill loss. These observations highlight the superiority of directly conditioning on skills rather than as auxiliary rewards in learning tasks with clearly distinguished skills, such as rearrangement.

\subsection{Robustness to Perturbations}
\label{sec:exp-robustness}

We conduct a perturbation test to evaluate the re-planning ability of \method and baselines in the \rhard task. The test perturbs the environment by closing the target receptacle after the agent had opened it for the first time. To succeed, the agent must detect the drawer has been closed and adjust the current plan to re-open the receptacle and retrieve the object. We compare \method's skill prediction and \mthree's task planner against ground truth skill labels derived from the privileged states of the simulation environment (detailed in \Cref{sec:gt-skill-label}). The test is performed on 400 episodes that involved only objects in the closed drawer. This test compares against the highest-performing modular baseline with access to oracle task plans, M3 (Oracle), and the highest-performing end-to-end baseline, DT (Skill). 

\begin{table}[h]
  \centering
\resizebox{0.99 \columnwidth}{!}{
\begin{tabular}{ccccc}
\toprule
& \multicolumn{2}{c}{Picking Target Object} & \multicolumn{2}{c}{Predicted Skill Accuracy} \\
\cmidrule(rl){2-3} \cmidrule(rl){4-5}
Method & \textbf{No Perturb} & \textbf{Perturb} & \textbf{No Perturb} & \textbf{Perturb} \\
\midrule
\textbf{M3 (Oracle) } &  24.00  &  0.50 & 44.00 &  10.00   \\
\textbf{DT (Skill)} &  10.00  &  8.50 & 39.00 &  33.99   \\
\rowcolor{NewBlue}
\textbf{ST (Ours)} &  \textbf{26.75}  &  \textbf{19.50} &  \textbf{60.00}  &  \textbf{47.00}   \\

\bottomrule
\end{tabular}
  }
  \caption{
    Success rates for picking up the target objects and accuracy of \Skillencoder module and \mthree's oracle-plan task planner under both perturbed and unperturbed environment, measured by ground truth skill labels. 
  }
  \label{tab:perturb-main}
  \vspace{-10pt}
\end{table}

The findings in \Cref{tab:perturb-main} demonstrate that \method outperforms the M3 (Oracle) baseline both in the success rate of picking up the object and skill selection accuracy. Specifically, when assessing the success rates of object retrieval, M3 (Oracle) performs comparably to \method in the absence of perturbations, but its performance significantly deteriorates to nearly 0\% with perturbations in the environment. In contrast, \method exhibits adaptability to changes in the state of the receptacle and the ability to re-open the receptacle, achieving a success rate for picking the object of 19.5\% in the presence of perturbations. Note that \method is not trained with any perturbations, and this behavior emerges from the design choices made in \method.

Similarly, the accuracy of the predicted skills demonstrates that \method outperforms M3 (Oracle) in high-level reasoning, both in the presence and absence of perturbations. Consistent with object retrieval, \method demonstrates a notably higher accuracy, indicating a higher level of adaptability to perturbations, compared to M3 (Oracle), which suffers from the non-reactive task planner that does not correctly respond to perturbations.

\begin{figure*}[t]
  \centering
  \begin{subfigure}[t]{0.28\textwidth}
    \includegraphics[width=\textwidth]{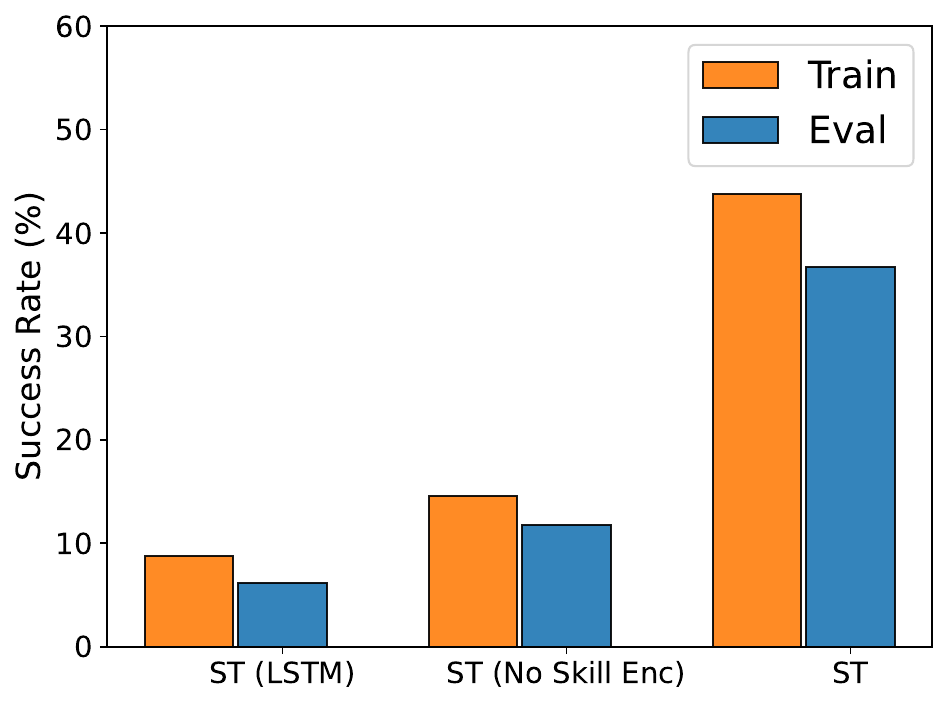}
  \vspace{-15pt}
    \caption{Architecture}
    \label{fig:arch}
  \end{subfigure}
  \begin{subfigure}[t]{0.28\textwidth}
    \includegraphics[width=\textwidth]{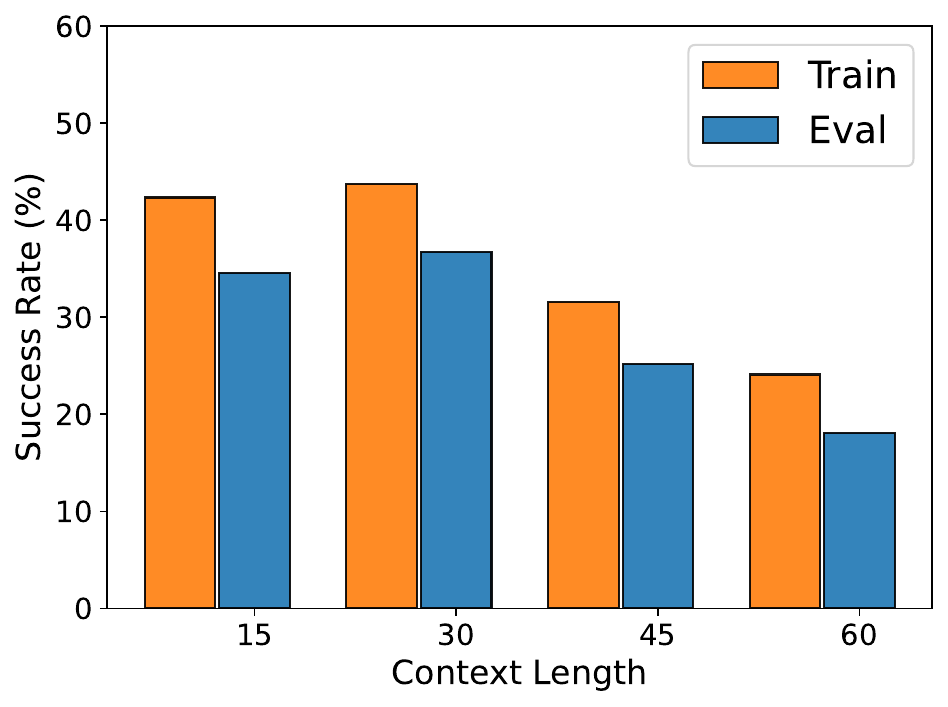}
  \vspace{-15pt}
    \caption{Context Length}
    \label{fig:contextlen}
  \end{subfigure}
  \begin{subfigure}[t]{0.28\textwidth}
    \includegraphics[width=\textwidth]{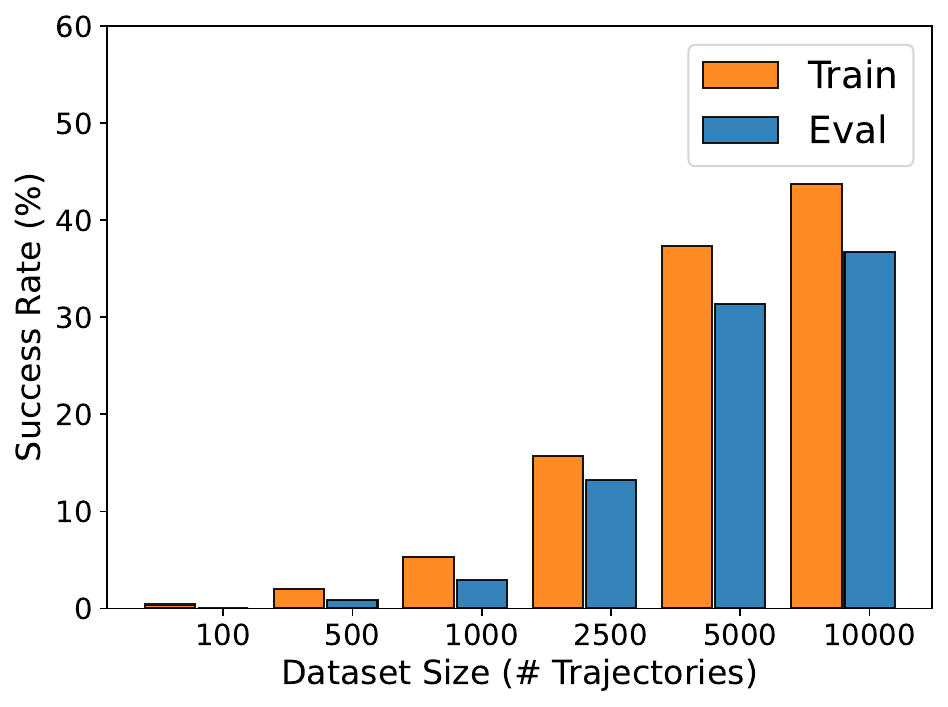}
  \vspace{-15pt}
    \caption{Dataset Size}
    \label{fig:dataset}
  \end{subfigure}
  \vspace{-5pt}
  \caption{
    Success rates for ablation and analysis experiments across different policy architectures (\cref{fig:arch}), context lengths (\cref{fig:contextlen}), and training dataset sizes (\cref{fig:dataset}) including performance on ``Eval" and ``Train" on the \emph{easy} split setting.
  }
\end{figure*}

We further compare \method to DT (Skill), another end-to-end method that has the potential to adapt to perturbations. \dtskill has a substantially lower success rate in picking the target object compared to \method, with less than half the success rate in both perturbed and unperturbed environments. In contrast to object retrieval, \dtskill displays only slightly lower accuracy in predicting correct skill labels than \method. Consistent with the qualitative findings in \Cref{sec:exp-rhard}, these observations suggest that although \dtskill demonstrates relatively good high-level reasoning ability, its primary challenge remains in learning and effectively leveraging distinct skills. On the other hand, we find that \method exhibits the combined advantages of adaptability to environment disturbances and proficiency in completing individual skills.

\subsection{Ablations and Analysis}
\label{sec:ablation}

We run ablations to address two key questions. Firstly, how necessary are the design choices of \method, such as the transformer backbone, the \Skillencoder module, and the choice of context length? Secondly, how do the dataset characteristics, such as its size and completeness, impact the performance? We perform all of the ablation experiments on the ``Easy" split of the task. 

\subsubsection{Policy Architecture Ablation}

In \Cref{fig:arch}, we evaluate the effectiveness of the transformer and the \Skillencoder module for mobile manipulation tasks. By replacing the Transformer backbone with an LSTM backbone, we observe a significant decline in performance. We find that the policy struggles to effectively execute individual skills, and we posit that the LSTM architecture is not as data-efficient for learning multiple skills. Our results also indicate a similarly low success rate when the \Skillencoder module is removed. Qualitatively, we discovered that without the \Skillencoder module, the policy is not able to differentiate between individual skills, and frequently executes inconsistent actions.

Context length refers to the maximum amount of historical information \method takes as input. 
We vary the context length from 15 to 60 timesteps and evaluate its impact on performance. Shown in \Cref{fig:contextlen}, a shorter context length of 15 timesteps performed slightly worse, while performance decreased as the context length increased from 30 to 60 timesteps. We posit that a shorter context length may not be able to fully capture the partially observable nature of the task, while a longer context length may lead to overfitting to the training dataset. Given the random robot start location during test time, the actual trajectory at test time will differ from that in the training dataset. The longer sequence exacerbates the discrepancy between the training dataset and the test time trajectory, resulting in an increased out-of-distribution error. A moderate context length may aid capturing short-term partially observed information and mitigating overfitting to the demonstration trajectories. The trend is similar to findings from previous works~\cite{zheng2022online, chen2021decision}.

\subsubsection{Dataset Size Analysis}

As previously demonstrated~\cite{brohan2022rt}, the size of the demonstration dataset is a vital aspect that impacts the efficacy of data-driven approaches such as \method. As shown in \Cref{fig:dataset}, we trained \method on subsets of the full demonstration dataset, ranging from 100 to 10k episodes. Our results showed that performance improved significantly when moving from 100 to 5k episodes, but the improvement is less pronounced when going from 5k to 10k demonstrations. 

\vspace{-5pt}
\subsubsection{Incomplete Demonstrations}
\vspace{-5pt}
In long-horizon tasks, obtaining sufficient demonstrations of the full trajectory can be challenging. We find that \method is also able to work effectively when trained with datasets containing only subsets of the full task. To demonstrate this, we train \method on only two subtasks of the task: navigate-pick and navigate-place, and it achieves a success rate of $36.5\%$ in ``Train" and $27.8\%$ in ``Eval" split, compared to $44\%$ and $37\%$ using complete trajectories. This ablation shows that, in contrast to \dt, which uses reward-conditioning across an entire episode, \method can also work with partial and incomplete demonstrations.

\section{Conclusion}
\label{sec:Conclusion}

This paper presents \method, a novel end-to-end method for embodied object rearrangement tasks. \method breaks the problem into predicting which skills to use and the low-level control conditioned on that skill, utilizing a causal transformer-based policy architecture. The proposed architecture outperforms baselines in challenging rearrangement problems requiring diverse skills and high-level reasoning over which skills to select. 

In future work, we will incorporate a more complex task specification, such as images of the desired state or natural language instructions. With the current geometric task specification, it is challenging to identify the target object after its container receptacle is manipulated. Specifically, because the object shifts inside a drawer when it is opened, its starting location no longer provides an indication of which objects is the target object. This limitation affects the success rate of retrieving objects in the challenging \rhard task.

\section{Acknowledgments}
The Georgia Tech effort was supported in part by NSF, ONR YIP, and ARO PECASE. The views and conclusions contained herein are those of the authors and should not be interpreted as necessarily representing the official policies or endorsements, either expressed or implied, of the U.S. Government, or any sponsor.

{\small
\bibliographystyle{ieee_fullname}
\bibliography{egbib}
}

\appendix

\section*{Appendix}
We include the source code at \href{https://bit.ly/3qH2QQK}{https://bit.ly/3qH2QQK}. We structure the Appendix as follows:

\begin{enumerate}[itemsep=2pt,topsep=0pt,parsep=0pt,partopsep=0pt,parsep=0pt,leftmargin=*]
    \item [\ref{sec:supp:method}] Additional details about the \method method.
    \item [\ref{sec:further-task-details}] Further description of the \reasy and \rhard tasks.
    \item [\ref{sec:exp-details}] Details about the demonstration dataset and baselines.
\end{enumerate}

\section{Further Method Details}
\label{sec:supp:method} 

For each training epoch of \method, we iterate over 80 randomly selected demonstrations. We step the learning rate schedule after each of these epochs. The learning rate schedule starts at a value of $ 1 \times 10^{-8}$ and then in a warmup phase, linearly increases this to a value of $ 6 \times 10^{-5}$ over 200 training epochs. The learning rate schedule then uses a cosine decay function that terminates at zero by the end of the training process. The method is trained until convergence, equating to iterating over the entire dataset approximately 6 times in total. 

In training \method for the \rhard task, we initialize the policy from the pre-trained policy that is trained only on 10,000 easy split episodes to speed up training times. Then, we include 10,000 demonstrations from the entire \rhard dataset to finetune the policy.

\section{Further Task Details}
\label{sec:further-task-details} 

The task settings for our experiments match those in the 2022 Habitat Rearrangement Challenge~\cite{habitatrearrangechallenge2022}. Our training episodes consist of 60 scene layouts from the ReplicaCAD dataset~\cite{szot2021habitat}. We use a subset of the 50,000 episodes from the Rearrangement Challenge as training episodes. Each episode has different object placements in the scene, and the robot is randomly spawned. We use a subset of these episodes to generate the 10,000 demonstrations for the imitation learning methods. 

 The agent controls the arm via delta joint position targets. The simulation runs PD torque control at 120Hz while the policy acts at 30Hz. The base is controlled through a desired linear and angular velocity. Like in~\cite{habitatrearrangechallenge2022}, we disabled the sliding behavior of the robot. This means that the robot will not move if its next base movement results in contact with scene obstacles. This modification significantly increases the difficulty of navigation in scenes with densely placed furniture. An object snaps to the robot's suction gripper if the tip of the suction gripper is in contact with the object and the grip action is active. 

The Fetch robot is equipped with an RGBD camera on the head, joint proprioceptive sensing, and base egomotion. The task is specified as the starting position of the object and the goal position relative to the robot's start in the scene. From these sensors, the task provides the following inputs:
\begin{itemize}[itemsep=0pt,topsep=0pt,parsep=0pt,partopsep=0pt,parsep=0pt,leftmargin=*]
  \item 256x256 90-degree FoV RGBD camera on the Fetch Robot head.
  \item The starting position of the object to rearrange relative to the robot's end-effector in cartesian coordinates.
  \item The goal position relative to the robot's end-effector in cartesian coordinates.
  \item The starting position of the object to rearrange relative to the robot's base in polar coordinates.
  \item The goal position relative to the robot's base in polar coordinates.
  \item The joint angles in radians of the seven joints on the Fetch arm.
  \item A binary indicator if the robot is holding an object (1 when holding an object, 0 otherwise). Note this provides information if the robot is holding any object, not just the target object.
\end{itemize}

The maximum episode horizon is 5,000 timesteps for the \rhard task. We increased the maximum number of timesteps for the task compared to~\cite{habitatrearrangechallenge2022} since we found this increased the success rate of all methods. 

We evaluate on 20 unseen scene configurations. We evaluate on a total of 400 episodes split into three difficulties as described in \Cref{sec:task}. Specifically, they consist of 100 (25\%) \emph{easy} episodes, 200 (25\%) of \emph{hard} episodes, and 100 (25\%) of \emph{very hard} episodes.

\section{Further Experiment Details}
\label{sec:exp-details} 

\begin{table*}[t]
  \centering
\resizebox{0.8 \textwidth}{!}{
  \begin{tabular}{ccccc}
\toprule
 & \textbf{DT / DT (Skill)} & \textbf{BC-Modular} & \textbf{ST (easy split)} & \textbf{ST (\rhard finetune)}  \\
\midrule
\textbf{Optimizer} &   AdamW  & AdamW & AdamW & AdamW   \\
\textbf{Learning Rate} & $ 6e^{-5}$ & $6e^{-5}$ & $6e^{-5}$ & $3e^{-5}$ \\
\textbf{Warm-Up Epochs} & 200 & 100 & 200 & 400 \\
\textbf{Episodes per Epoch} & 80 & 80 & 80 & 80 \\
\textbf{Total Epochs} & 750 & 300 & 750 & 600 \\
\bottomrule
\end{tabular}

  }
  \caption{
    Hyperparameters for all imitation learning based methods. All methods are trained with the same dataset.
  }
  \label{tab:il_hyperparams}
\end{table*}

\begin{table}[t]
  \centering
\resizebox{0.8 \columnwidth}{!}{
  \begin{tabular}{cccc}
\toprule
 & \textbf{TP-SRL} & \textbf{M3} & \textbf{Mono} \\
\midrule
\textbf{Optimizer} &  Adam  &  Adam & Adam \\
\textbf{Learning Rate} &  $ 3e^{-4}$ & $ 3e^{-4}$ & $ 2.5e^{-4}$ \\
\textbf{\# Steps Per Policy} &  100M & 100M & 100M \\
\textbf{Value Loss Coef} &  $ 0.5$ & $0.5$ & $0.5$ \\
\bottomrule
\end{tabular}

  }
  \caption{
    Hyperparameters for all reinforcement learning based methods. All these methods are trained with PPO. For TP-SRL and M3, these hyperparameters are used for training all the skill policies.
  }
  \label{tab:rl_hyperparams}
\end{table}

\subsection{Data Generating Policy Details}
\label{sec:expert-details} 
As described in \Cref{sec:exp-setup}, we use \mthreeo as our data-generating policy. We follow the same setup as~\cite{gu2022multi} for training this policy. Specifically, we first train mobile manipulation policies including navigation, pick, place, open drawer, and open fridge policies. Each policy is trained for 100M steps using PPO. We use the same reward functions as from~\cite{gu2022multi}. \mthreeo then composes these skills together using an oracle planning module. Using the ground truth environment state, the oracle planner detects if the object or goal is in a closed receptacle and then adapts the plan accordingly. 

We use \mthreeo to generate a dataset of demonstration trajectories for the \rhard task. We run \mthreeo on the Habitat Rearrangement Challenge train dataset~\cite{habitatrearrangechallenge2022}. We record the observations, actions, rewards, and which skill is currently executing. The rewards are used for training \dt and \dtskill. The skill labels are used for training ST and \dtskill.

\subsection{Baseline Details}
\label{sec:baseline-details} 
In this section we provide details on the baselines in \Cref{sec:exp-baseline}:

The monolithic RL (\mono) baseline follows the monolithic RL baseline implementation from~\cite{szot2021habitat}. \mono inputs the visual representation into an LSTM network and then inputs the recurrent state into an actor and value function head. \mono is trained with PPO for 100M steps across 4 GPUs with 32 environment workers per-GPU, taking 21 hours to train. We fix the parameter count of \mono and all other end-to-end baselines to be the same as \method.

For \dt and \dtskill, we follow the details from~\cite{chen2021decision}. Note that both \dt and \dtskill are conditioned on the desired \emph{return-to-go}, which is the sum of future rewards. We use the \emph{move-object} reward for the entire \rhard task to train these methods. This rewards the progress of the robot to move the target object from its start position to the goal position. \dtskill is also trained with an auxiliary classification objective for which skill is currently active. At evaluation time, we condition on the maximum return from the demonstrations. \dt and \dtskill both require access to the reward function at inference time, whereas \method does not. 

We train \mthree and \mthreeo using the process described in \Cref{sec:expert-details}. For \mthree, we execute the fixed task plan of navigating to the object, picking up the object, navigating to the goal, and then placing the object at the goal. Note that \mthreeo only adjusts its initial task plan based on the starting environment state. It does not dynamically adjust its task plan. For example, if the agent fails to open the drawer correctly, it won't retry the open skill but instead continues with the pick skill.
Like in~\cite{szot2021habitat}, TP-SRL uses the same training process as M3 but does not include base movement for the pick and place skills. The navigation skill is also rewarded for moving to the closest navigable point to its goal rather than a nearby region as with M3. 

The \bcmodular methods train the individual skills with imitation learning using the same demonstration dataset as \method. These methods are deployed in the same fashion as \mthree and \mthreeo.

Hyperparameters for all RL methods are detailed in \Cref{tab:rl_hyperparams} and all imitation learning methods in \Cref{tab:il_hyperparams}. The set of hyperparameters is the same across both tasks unless stated otherwise in the table.

\subsection{Ground Truth Skill Labeling Rules}
\label{sec:gt-skill-label} 
The rules used to define the ground truth skill labels are the following: 
\begin{itemize}[itemsep=0pt,topsep=0pt,parsep=0pt,partopsep=0pt,parsep=0pt,leftmargin=*]
\item \textbf{Navigate} if the agent is not within a threshold of 1.5m from either the object start or the goal location. We used 1.5m to be consistent with \mthreeo's training setting. 
\item \textbf{Pick} if the agent is within 1.5m from the object start location and either the object is not hidden in a receptacle or the receptacle containing the object is already opened. 
\item \textbf{Place} if the agent is within 1.5m from the object goal location and either the goal is not hidden in a receptacle or the receptacle containing the goal is already opened. 
\item \textbf{Open-Receptacle} if the agent is within 1.5m from either the object start or the goal location and the receptacle containing the object or the goal is not opened. 
\end{itemize}

\end{document}